\documentclass[10pt, a4paper]{article}

\usepackage{lrec-coling2024}

\usepackage{subfig}
\usepackage{graphicx}
\usepackage[utf8]{inputenc}
\usepackage{booktabs}
\usepackage{multirow}
\usepackage{microtype}
\usepackage{tikz-dependency}
\usepackage{microtype}
\usepackage{graphicx}
\usepackage{booktabs}
\usepackage{xcolor,pifont}
\usepackage{float}
\usepackage{comment}
\usepackage{footnote}
\usepackage{appendix}
\usepackage{tabularx}
\usepackage{tabularx, booktabs}
\PassOptionsToPackage{hyphens}{url}\usepackage{hyperref}
\usepackage{xurl}
\usepackage{enumitem}

\title{A Tulu Resource for Machine Translation} 
\name{Manu Narayanan, Noëmi Aepli} 

\address{University of Zurich \\
         \{manu.narayanan,noemi.aepli\}@uzh.ch\\
        }

\abstract{
We present the first parallel dataset for English--Tulu translation. Tulu, classified within the South Dravidian linguistic family branch, is predominantly spoken by approximately 2.5 million individuals in southwestern India. Our dataset is constructed by integrating human translations into the multilingual machine translation resource FLORES-200.
Furthermore, we use this dataset for evaluation purposes in developing our English--Tulu machine translation model. For the model's training, we leverage resources available for related South Dravidian languages. We adopt a transfer learning approach that exploits similarities between high-resource and low-resource languages. This method enables the training of a machine translation system even in the absence of parallel data between the source and target language, thereby overcoming a significant obstacle in machine translation development for low-resource languages.
Our English--Tulu system, trained without using parallel English--Tulu data, outperforms Google Translate by 19 BLEU points (in September 2023). 
The dataset and code are available here: \url{https://github.com/manunarayanan/Tulu-NMT}.
 \\ \newline  }
 
\begin{document}

\maketitleabstract
\section{Introduction}

Over the past decade, the field of neural machine translation (NMT) has seen significant advances with the advent of sequence-to-sequence models~\citep{seq2seq}, attention mechanisms~\citep{bahdanau}, and transformer architecture~\citep{attention}. 
However, these advancements fall short when confronted with languages lacking extensive parallel datasets. Challenges stemming from both the scarcity of abundant parallel data and the absence of domain-diverse data pose significant hurdles in crafting robust NMT models \citep{6challenges}.
Regrettably, a vast majority of the world's linguistic diversity, spanning over 7,000 languages, faces one or both of these challenges \citep{littauer2016open, Lakew2020LowRN}.

Among these languages stands Tulu (ISO 639-3 code: TCY), a South-Dravidian language spoken by approximately 2.5 million individuals in India \citeplanguageresource{madasamy-etal-2022-overview-this}, characterized by several dialects \citep{ethnologue}. 
Tulu is not recognized as an official language, neither in India nor in any other country. Hence, it is not used for official purposes and education, where Kannada or Malayalam is used instead.
However, efforts to enhance accessibility to the language have been evident through Unicode proposals for a Tulu script and petitions urging the Indian government to recognize Tulu as an official state language \citep{hindustan_times_2023}.
Furthermore, Tulu demonstrates a notable online presence and engagement among its speakers through various social media platforms. For instance, \textit{Jai Tulunad}\footnote{\url{https://jaitulunad.in/}}, a volunteer organization established in 2014 in Karnataka, India, maintains active profiles on several social media platforms with over 1,000 engaged subscribers. They launched an online English-Tulu dictionary in 2021 and regularly host cultural and educational events for Tulu speakers. 
Furthermore, several groups on social media are exclusively dedicated to Tulu language memes and other content.\footnote{\url{https://www.facebook.com/VenurTroll}} Moreover, Tulu features a vibrant film industry, which produced nine movies in 2023.\footnote{\url{https://en.wikipedia.org/wiki/List_of_Tulu_films_of_2023}}

\begin{figure}
    \centering
    \includegraphics[width=0.4\textwidth]{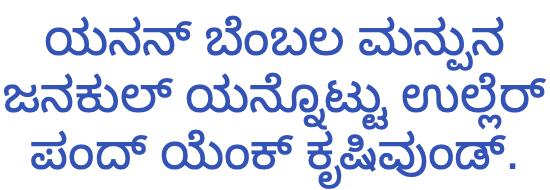}
    \caption{A sentence in Tulu taken from our human-translated extension of the FLORES-200 dataset. English: `\textit{I am happy that there are people willing to support me.}'}
    \label{fig:enter-label}
\end{figure}

Considering the surge of information driven by the internet and social media, coupled with the increasing importance of language accessibility in contemporary society, it becomes crucial to create resources and methodologies that tackle the translation challenges encountered by low-resource languages. This endeavor plays a crucial role in promoting social equity, economic equality, and political inclusivity.

Transfer learning offers an approach to mitigate the low-resource issue to a certain extent. It aims to use existing knowledge to adapt pre-trained models. Instead of starting from scratch, this technique facilitates the adaptation of already-trained models to new languages, particularly when a related language with such resources is available.
Tulu is fortunate in this regard, as Kannada (ISO 639-3 code: KAN) serves as a closely-related language with some readily available NLP resources.

In this study, we present the first parallel dataset for Tulu and use it to evaluate our machine translation system for English--Tulu. Without access to parallel EN--TCY data, we developed this system using a transfer learning \citep{zoph-etal-2016-transfer} to address translation challenges in this low-resource language.
Our main contributions are:
\begin{itemize}
    \item Introducing a machine translation dataset for Tulu by extending FLORES-200 with human translations into Tulu.
    \item Developing a machine translation system for English--Tulu, leveraging the resources of related Dravidian languages and employing transfer learning.
\end{itemize}

\section{Linguistic Context} \label{linguistic}

Languages spoken in the South Asian region belong to at least four major language families: Dravidian, Austro-Asiatic, Sino-Tibetan, and Indo-European (predominantly from the Indo-Aryan sub-branch). Among these, the Dravidian languages constitute the second-largest group.\footnote{See \url{https://censusindia.gov.in/nada/index.php/catalog/42561}, page 14.}

The Dravidian language family ranks as the fifth largest language family globally, comprising approximately 25 languages primarily spoken in India \citep{SUBRAHMANYAM2006785}. This family is divided into four subgroups: North Dravidian, South Dravidian, South-Central Dravidian, and Central Dravidian. Tulu, Kannada, and Malayalam belong to the South Dravidian subgroup.

Dravidian languages generally share the following main characteristics:\\
\textbf{Vowels} Most of the Dravidian languages have `a 10-vowel system, with five short and five long ones'~\citep{SUBRAHMANYAM2006785}.\\
\textbf{Consonants} Retroflex consonants, a distinctive feature rare outside the Indian subcontinent, are prominent in Dravidian languages. However, voiced stops and aspirated stops are notably absent in these languages.\\
\textbf{Cases} According to~\citep{steever_2017}, Dravidian languages typically feature between five and eight cases. These include nominative, accusative, dative, genitive, locative (`in'), ablative (`from'), sociative (`with'), and instrumental (`by'). Kannada and Tulu exhibit all eight cases, while Malayalam has seven, with the absence of the ablative case.\\
\textbf{Morphology} Dravidian languages are characterized as agglutinative, with grammatical relations such as voice or tense typically expressed through suffixation and compounding.\\
\textbf{Syntax} Word order is a flexible subject-object-verb, with the verb always in the final position.\\
\textbf{Writing} The primary Dravidian scripts in current use include Kannada, Malayalam, Tamil, and Telugu. The Tigalari script, historically used for writing Tulu, has gradually fallen out of use over the past few centuries, leading to the adoption of the Kannada script for writing Tulu~\citep{steever2019dravidian}.

\paragraph{Kannada} is spoken by approximately 43.7 million people in India, according to \citet{language_atlas_kan}, with around 93\% of them residing in Karnataka, where it holds the status of the official language.  
Kannada shares most of the typical characteristics of the Dravidian languages listed above.

Tulu and Kannada have co-evolved in close geographical and cultural proximity since at least the 8th century CE. Since 1947, Kannada has served as the official language in the Tulu-speaking region, with the exception of Kasaragod district, where Malayalam holds official status. Consequently, there has been an increasing trend of using Kannada loanwords in contemporary Tulu. Numerous Tulu words exhibit notable similarities to their Kannada equivalents. 
Hence, Kannada resources can serve as a starting point for constructing NLP systems tailored to Tulu.

\paragraph{Tulu}
is spoken by around 2.5 million people~\citeplanguageresource{madasamy-etal-2022-overview-this} in the Dakshina Kannada and Udupi districts of Karnataka state and the Kasargod district of the Kerala state, with scattered speakers found in Maharashtra and other regions of India. Malayalam and Kannada share linguistic ties with Tulu, which comprises several dialects \citep{ethnologue}.
The benchmark dataset introduced in this study is closely aligned with Central Tulu, which is predominantly spoken in the Mangaluru region and serves as the primary city in the Tulu-speaking area. We employ the Kannada script for written communication, reflecting the prevailing practice among Tulu speakers.

The grammatical aspects of Tulu are similar to other South Dravidian languages~\citep{brigel1982grammar}. The majority of Tulu speakers are bilingual, often using Kannada or Malayalam when communicating with individuals outside their community within their respective states. The Tigalari script, traditionally used for writing Tulu, has been gradually replaced by Kannada. A significant contributing factor to this transition is the absence of a Unicode script that supports Tigalari characters. Additionally, the Tulu community has refrained from writing in Tulu for several generations, opting to use Kannada or Malayalam for official purposes and education. Consequently, the Tigalari script has fallen out of use over the generations.

\section{Related Work}

The \textit{Shared Task on Translation of Under-Resourced Dravidian Languages} at the \textit{DravidianLangTech-2022} workshop~\citeplanguageresource{madasamy-etal-2022-overview-this} involved Kannada--Tulu as one of the language pairs to be translated. The participants were given a Kannada-Tulu parallel training dataset of 8,300 sentences and development and test sets containing 1,000 sentences each. These datasets were created by collecting monolingual Tulu documents from digitally accessible sources and manually translating them into Kannada. The team that scored the highest BLEU score \citep{papineni-etal-2002-bleu} for Kannada--Tulu translation trained a transformer model provided by OpenNMT \citep{klein-etal-2017-opennmt} 
for five different Dravidian languages (Tamil, Malayalam, Telugu, Kannada and Tulu) and got a BLEU score of 61.49~\citep{goyal-etal-2022-translation}. They trained their model for Tulu using only the 8,300 sentences of the training set. They hypothesize that the BLEU score might be higher because of the similarity of training and test sentences and, thus, high word overlap in the source and target data. A word or even sentence overlap in source and target might occur since Kannada and Tulu are similar with quite some shared vocabulary.

\citet{10.1145/3587932} focus specifically on translating low-resource Indic languages by developing a multilingual NMT system with a shared encoder-decoder containing 15 language pairs (i.e. English and 14 Indic languages). They utilize the similarity between Indic languages, along with back-translation and domain adaptation, to achieve better results in translating low-resource languages. However, all the language pairs explored in the paper have parallel training data available, and back-translation is used only to create synthetic parallel training data from monolingual sentences.

The transfer learning approach NMT-Adapt~\citep{ko-etal-2021-adapting} aims to leverage the lexical and syntactic structure similarities between high-resource languages and low-resource languages and train a translation model without using any parallel data. It combines denoising autoencoding~\citep{artetxe2018unsupervised}, back-translation~\citep{sennrich-etal-2016-improving}, and adversarial objectives to utilize monolingual data for low-resource adaptation. \citet{ko-etal-2021-adapting} experimented with three groups of languages, namely Iberian languages, Indic languages, and Arabic. Within the Indic languages, they treated Hindi as the high-resource language and Marathi, Nepali, and Urdu as related low-resource languages.
\citet{sennrich-etal-2016-improving} had already shown that pairing monolingual data with automatic back-translation and using that as additional parallel training data can improve the capability of an English--\textit{German} model to translate \textit{Turkish}--English. \citet{ko-etal-2021-adapting} added denoising autoencoding to this task, such that the model learns a shared feature space for the high-resource and low-resource languages and enables the encoder and decoder to transform between the features and the sentences. This involves adding noise to datasets in both languages by randomly shuffling words by at most 3-word positions and masking words with a uniform probability of 0.1. A dataset with this `noised' data, along with the original data, is used to do the denoising autoencoding. The denoising autoencoding trains the model to reconstruct the original version of a corrupted input sentence~\citep{artetxe2018unsupervised}. According to \citet{lample2018unsupervised}, this enables the feature space to learn high-level semantic knowledge and make it more robust.

\section{The 1\textsuperscript{st} Dataset for Tulu MT}\label{sec:dataset}

To maximize the potential impact, we opted to extend an already existing and widely adopted benchmark dataset for the creation of a Tulu dataset: \textit{FLORES-200}, `Evaluation Benchmark for Low-Resource and Multilingual Machine Translation'\footnote{`The sentences were sampled in equal amounts from  Wikinews (an international news source), Wikijunior (a collection of age-appropriate non-fiction books), and Wikivoyage (a travel guide).' \url{https://github.com/openlanguagedata/flores}} \citeplanguageresource{flores,nllb-22}, which contains over 200 language varieties to-date.
The FLORES-200 dataset comprises 2009 sentences for each language.

To obtain the translations, we collaborated with the organization \textit{Jai Tulunad}\footnote{\url{https://www.jaitulunad.com/about}}, a volunteer organization headquartered in the southern Indian city of Mangaluru. 
This collaboration was crucial not only for us to find native Tulu speakers but also because an increasing number of researchers have underscored the significance for NLP researchers to prioritize the needs and preferences of the pertinent speaker community \citep{bird-2020-decolonising,bird-2022-local,liu-etal-2022-always,mukhija2021designing,blaschke2024dialect}. The organization and the translators were happy to contribute to our project
which makes us confident that this is in the interest of the Tulu community.
\textit{Jai Tulunad} is dedicated to preserving and promoting Tulu language and culture. Approximately 15 volunteers from this organization, based in Mangaluru, Karnataka state, India, participated in translating the sentences from the FLORES-200 dataset into Tulu. While the volunteers are not professional translators, they are all native Tulu speakers fluent in English. Furthermore, they all have native proficiency in Kannada since Kannada is the language taught in schools, used for official purposes and for communicating with people in non-Tulu speaking communities in the same region.
Among the translators, two are Tulu language instructors, and one is a distinguished Tulu poet.

During the translation process, translators referred to both English and Kannada sentences in the dataset and translated them into Tulu. They consulted with literary experts within the translator group to address questions regarding Tulu vocabulary and resolved decisions regarding the utilization of outdated Tulu words, colloquialisms, and loanwords from Kannada in the translation.
Along with the original FLORES-200 dataset, we provided the translators with guidelines, which we adapted from the original guidelines that were published by \citeplanguageresource{flores,nllb-22}.

\begin{enumerate}
    \item Translations must be neutral, informative, and clear to native speakers.
    \item No assistance from any machine translation tools; this was easy since there are no existing machine translation tools for Tulu.
    \item Proper nouns may be transliterated if no equivalent term exists in Tulu. Similarly, abbreviations must be translated in the manner they usually appear in Tulu.
    \item Idioms, metaphors, etc.\ need not be translated word by word. They should be translated as they usually appear in Tulu.
    \item The most knowledgeable individuals among the team (in this case, the Tulu school teachers and the poet) shall take on the role of experts, who will resolve any queries that the other translators may have. They shall also review all the sentences translated by everyone, including themselves, to eliminate typos, grammatical errors, and any other translation errors.
\end{enumerate}

\begin{figure}
    \centering
    \includegraphics[width=0.4\textwidth]{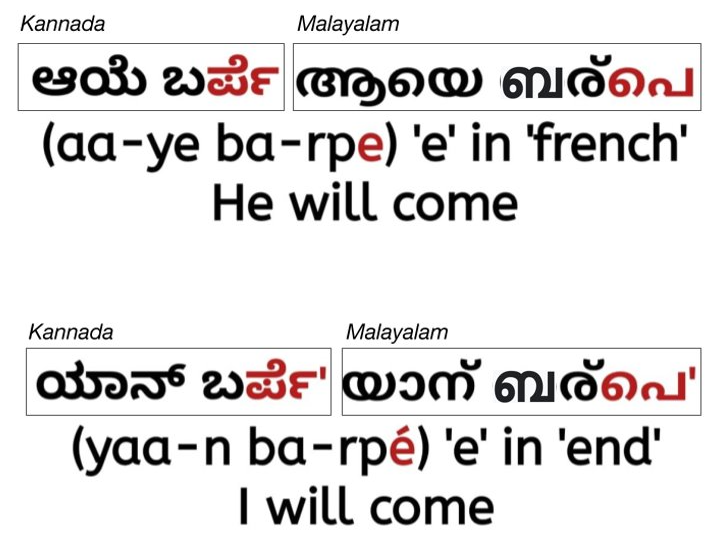}
    \caption{Example sentences showing the two \texttt{/e/} sounds in Tulu. Image shared by the translation team of \textit{Jai Tulunad}.}
    \label{fig:e_sample}
\end{figure}

Throughout the translation process, we had regular conversations with two members of the team who coordinated the whole process. The major challenges the translators faced during this process are listed below:

\begin{enumerate}
    \item While the vocabulary of Tulu historically encompassed a vast array of words, many of them have fallen out of common use today, particularly as they are not taught in schools as part of contemporary Tulu. Additionally, many Tulu words have been replaced by Kannada or Sanskrit\footnote{\url{https://www.ethnologue.com/language/san/}} equivalents, further complicating the preservation of the language. As a result, the team encountered the need for frequent considerations to determine the appropriate word choices in specific instances, given the absence of a widely accepted standard Tulu.
    \item Passive voice is not commonly used in Tulu, nor in any other modern Dravidian languages according to \citet{krishnamurti_2003}. Consequently, a literal translation of an English sentence in passive voice may sound unusual in Tulu. Hence, such sentences were translated into the commonly used active voice.
    \item Dialectical variations exist within Tulu based on the region of origin of the speaker. Therefore, the team had to pay attention to using a consistent dialect. In this case, all translators adhered to the Mangaluru (Central Tulunad) dialect.
    \item There are two variations of the phonetic \texttt{/e/}
    sound (a close-mid front unrounded vowel) in Tulu, which occur when it is in the word-final position, a unique feature of Tulu~\citep{SUBRAHMANYAM2006785}. However, these distinctions cannot be represented in the English, Kannada, or Malayalam scripts. The proposal for a Unicode Tulu script, that was submitted by members of \textit{Jai Tulunad} to the Unicode Consortium,\footnote{\url{https://unicode.org/consortium/consort.html}} addresses these differences. An example usage of the two \texttt{/e/} sounds is shown in Figure \ref{fig:e_sample} 
    It displays the Tulu translation of the sentences \textit{`\textbf{He} will come'} and \textit{`\textbf{I} will come'} written using Kannada and Malayalam scripts, with an apostrophe (\texttt{'}) used to differentiate the two \texttt{/e/} sounds.
    \item Another challenge, which is also prevalent in other language pairs, is translating the word \textit{you} into Tulu. Whereas English has only one \textit{you} and Kannada has two (singular \texttt{neenu} and plural \texttt{neevu}), Tulu has three words for \textit{you}: singular \texttt{ee}, plural \texttt{nikulu} and formal \texttt{eeru}. As a result, there could be some ambiguity regarding the specific Tulu word for \textit{you} when translating standalone English sentences.  
\end{enumerate}

\section{Experiments}

\subsection{Data}

\textbf{English--Kannada training set}
To train the base model 
for EN--KN translation, we use the Samanantar dataset~\citeplanguageresource{Samanantar}, comprising 4 million sentences. This dataset primarily originates from Indian websites, government documents, and sources such as Coursera, Khan Academy, and select science YouTube channels,\footnote{\url{https://www.coursera.org} \& \url{https://www.khanacademy.org} \& \url{https://www.youtube.com}} which offer parallel human-translated subtitles in various Indic languages~\citeplanguageresource{Samanantar}. 

\textbf{English--Kannada test set}
To test the EN--KN models, we use the \texttt{dev} and \texttt{dev-test} sets of the FLORES-200~\citeplanguageresource{flores,nllb-22} dataset, consisting of 997 and 1,012 sentences, respectively. 
The domain mainly uses Wikimedia sources such as WikiNews, WikiJunior, and WikiVoyage. This stands in contrast to the training data from Samanantar, which is primarily derived from Indian sources situated within the Indian context.

\textbf{English--Tulu test set}
To evaluate our EN--TCY models, we use our newly developed dataset introduced in Section \ref{sec:dataset}, which comprises a set of 1,300 human-translated sentences from FLORES-200. We split this dataset into 647 sentences for the development set and 653 for the test set.\footnote{When we conducted the experiments, not all of the 2,009 sentences have been translated.}

\textbf{Monolingual Tulu dataset}
The method we apply requires a monolingual dataset in the low-resource language for the back-translation and denoising autoencoding steps. As there is no pre-existing dataset readily accessible for Tulu, we have turned to the Tulu Wikipedia\footnote{We downloaded the Tulu Wikipedia (\url{https://tcy.wikipedia.org}) articles via Wikimedia Downloads (\url{https://dumps.wikimedia.org/}) and used the WikiExtractor \citep{Wikiextractor2015} to get the text.} containing 1,894 articles~\citep{Tulu_Wikipedia}. 
As a result of processing the articles, we obtained a monolingual Tulu corpus comprising 40,000 sentences.

\begin{table}
\centering
\small
\begin{tabular}{ll|r}
\bf{Dataset}   & \bf{Source}   & \bf{\#sents}    \\
\midrule
EN--KN training   & Samanantar & 4,093,524    \\
EN--KN test & FLORES-200    & 2,009 \\
TCY monolingual    & Wikipedia   & 40,124    \\
EN--TCY test   & Human transl. FLORES& 1,300 \\
EN--TCY training    & DravidianLangTech-22  & 8,300
\end{tabular}
\caption{Datasets used in our experiments.}
\end{table}

\begin{figure}
    \centering
    \includegraphics[width=0.4\textwidth]{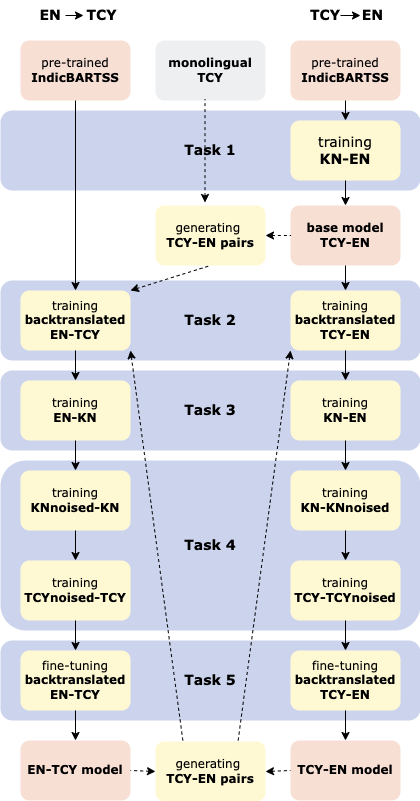}
    \caption{Experimental setup illustrating the steps involved in the training.}
    \label{fig:method}
\end{figure}

\subsection{IndicBARTSS \& YANMTT}

The original version of the method we use, NMT-Adapt~\citep{ko-etal-2021-adapting}, is based on the multilingual BART (mBART) language model~\citep{mbart} to initialize training. However, Kannada is not included neither in mBART nor mBART-50. Consequently, Tulu data would need to be transliterated into one of the languages included in mBART-50, such as Malayalam, which would affect the performance of the model.

Therefore, we opted for IndicBARTSS\footnote{\url{https://huggingface.co/ai4bharat/IndicBARTSS}}, an updated version of IndicBART~\citep{indicbart}.
IndicBARTSS supports eleven Indic languages in their native scripts, including Kannada. 
It is a multilingual model trained on the IndicCorp corpus, which was introduced by \citet{indicnlpsuite}.
IndicCorp is a collection of monolingual corpora in eleven Indic languages and English. It contains 452 million sentences (five billion tokens). These texts were crawled from online sources, primarily comprising news articles, magazines, and books.
Additionally, \citet{dabre-etal-2023-yanmtt} developed the YANMTT toolkit, which is independently maintained by some of the researchers associated with IndicBARTSS. We use this toolkit in our experiments for pre-training, fine-tuning, and decoding. 

\subsection{NMT-Adapt}

We adopted a slightly simplified version of the transfer learning approach NMT-Adapt introduced by \citet{ko-etal-2021-adapting} to develop a machine translation system for English--Tulu. This approach combines denoising autoencoding \citep{artetxe2018unsupervised} and back-translation \citep{sennrich-etal-2016-improving} into a multi-step, iterative procedure, as depicted in Figure \ref{fig:method}. While striving to closely follow the original implementation, we omitted an adversarial step due to limitations of the pre-trained model and training library we utilized (see Section \ref{limitations}). Our aim was to replicate and compare the effectiveness of this process for our set of languages wherever possible (see Section \ref{limitations}).

\textbf{Task 1: Fine-tuning for back-translation}
The initial step involved fine-tuning the IndicBARTSS model to translate from Kannada to English. To achieve this, we used the Samanantar dataset for training and the EN--KN dataset from FLORES-200 for development and testing. Subsequently, we used this model to translate the monolingual Tulu dataset into English. This fine-tuned model serves as the base model for TCY--EN translation.

\textbf{Task 2: Training with back-translation}

For the training with back-translation, we used a second pre-trained IndicBARTSS model, which served as the base model for EN--TCY translation. We trained this model using the back-translation pairs obtained in Task 1.

\textbf{Task 3: Training with parallel data}

In the third step, we trained the model from the previous task with the parallel English--Kannada training dataset from Samanantar.

\textbf{Task 4: Denoising autoencoding}

For the denoising autoencoding, we generated `noised' sentences from the monolingual Tulu and Kannada training datasets by implementing random shuffling and word masking. We followed the method described in \citet{ko-etal-2021-adapting}, where words are randomly shuffled with a maximum shift of three positions, and each word is masked with a uniform probability of 0.1. 
Subsequently, we trained the English--Tulu base model using these `noised' Tulu and Kannada sentences as the source data, with the unaltered Tulu and Kannada sentences as the target data.

\textbf{Task 5: Fine-tuning with back-translation}

In the final step, we further fine-tuned the EN--TCY model from the preceding step using the back-translated pairs. We used this model to generate the English side of the back-translated pairs again, forming the new back-translation set for the next step. 
These newly generated back-translated pairs are used to repeat Tasks 2--5 on the TCY--EN base model obtained in Task 1. This process could then be repeated by each model supplying the back-translation pairs for the other one.

\subsection{Fine-tuning with EN--TCY data}

\citetlanguageresource{madasamy-etal-2022-overview-this} released parallel training data for KN--TCY, comprising 8,300 sentences as part of the \textit{Shared Task on Translation of Under-Resourced Dravidian Languages} at the \textit{DravidianLangTech-2022} workshop.
We used this data to create a parallel EN--TCY dataset and fine-tuned the models obtained from the modified NMT-Adapt process we followed.
To create the parallel EN--TCY dataset from KN--TCY sentences, we initially translated the Kannada sentences into English. We performed this translation using two methods: first, utilizing the `base model TCY--EN' obtained in Task 1, and second, using Google Translate\footnote{\url{https://translate.google.com}} for comparison. 
Note that Google Translate does not currently offer a translation service for Tulu. However, it does support Kannada translation. Given the shared script and substantial linguistic similarity between Tulu and Kannada, as described in Section \ref{linguistic}, we opted to utilize Google Translate to translate Tulu text as if it were Kannada. Despite the absence of alternative online translation services for Tulu, we believe this approach serves as a meaningful and comprehensible starting point for benchmarking the performance of our model.

Although the `base model TCY--EN' achieved a BLEU score \citep{papineni-etal-2002-bleu} of 30.65 when evaluated on the FLORES-200 Kannada \texttt{dev-test} split, it showed lower translation quality compared to Google Translate, as it omitted some information from the source sentence during translation.
Therefore, we combined the English sentences translated by Google Translate with the Tulu sentences from the KN--TCY dataset to obtain an EN--TCY parallel dataset comprising 8,300 sentences. 
Finally, we fine-tuned the models obtained at the end of Task 5 in both the EN--TCY and TCY--EN directions using this dataset. Our aim was to assess whether we could further enhance the models beyond what NMT-Adapt offers.

\section{Results}

We evaluated the machine translation model and each task of the process with our new test set as introduced in Section \ref{sec:dataset}, using SacreBLEU\footnote{\url{https://github.com/mjpost/sacrebleu}}~\citep{sacrebleu}.
Table \ref{tab:results} presents the BLEU scores for each stage of the training process. 
The TCY--EN model, obtained by fine-tuning the pre-trained IndicBARTSS on the Samanantar EN--KN data, achieves a BLEU score of 1.84, suggesting that the model is not capable of translating Tulu.

\begin{table*}[ht]
\centering
\begin{tabular}{c|cclc|r}
\multicolumn{1}{l}{\textbf{Iteration}} & \multicolumn{1}{l}{\textbf{Direction}} & \textbf{Task no.} & \textbf{Task} &  \textbf{Lannguages} & \textbf{BLEU} \\ \midrule
\multirow{10}{*}{1} & TCY–EN &  1 & fine-tuning with & KN–EN & 1.84 \\
 & EN–TCY & 2 & back-translation with & EN–TCY & 12.83 \\
 & EN–TCY & 3 & training with parallel & EN–KN & 17.27 \\
 & EN–TCY & 4a & denoising autoencoding with & KN & 3.20 \\
 & EN–TCY & 4b & denoising autoencoding with & TCY & 5.92 \\
 & EN–TCY & 5 & fine-tuning with back-translation data &  & 11.06 \\
 & TCY–EN & 2 & back-translation with & TCY–EN & 19.53 \\
 & TCY–EN & 4a & denoising autoencoding with & KN & 7.08 \\
 & TCY–EN & 4b & denoising autoencoding with & TCY & 7.08 \\
 & TCY-EN & 5 & fine-tuning with back-translation data &  & \textbf{25.97} \\ \midrule
\multirow{5}{*}{2} & EN–TCY & 2  & back-translation with & EN–TCY & 12.09 \\
 & EN–TCY & 3 & training with & EN–KN & 9.09 \\
 & EN–TCY & 4a & denoising autoencoding with & KN & 3.45 \\
 & EN–TCY & 4b & denoising autoencoding with & TCY & 6.59 \\
 & EN–TCY & 5 & fine-tuning with back-translation data &  & 13.43
\end{tabular}
\caption{BLEU scores for each step in the training with 2 iterations.}
\label{tab:results}
\end{table*}

\paragraph{English-Tulu Translation}

The fine-tuned TCY--EN model from Task 1 was used to translate the monolingual Tulu data into English. Subsequently, the sentence pairs obtained from the EN--TCY translation were used to fine-tune the IndicBARTSS model, which served as the base model for translating from English to Tulu. This resulted in a BLEU score of 12.83 for Tulu. While this score signifies a certain level of learning, the translation cannot be considered useful.

Moving on to Task 3, we trained the EN--TCY model using the Samanantar EN--KN data. This task improved the BLEU score for Tulu, reaching 17.27. This enhancement suggests that training the model to translate into Kannada also enhanced its ability to translate into Tulu. 
This improvement can be attributed to the high degree of similarity between Tulu and Kannada and the effectiveness of transfer learning. 

In Task 4a, after conducting denoising autoencoding with Kannada data, the BLEU score decreased to 3.20. However, it then increased slightly to 5.92 upon denoising autoencoding with Tulu in Task 4b.

The adversarial training task, a component of the original NMT-Adapt method, was not implemented in this work. The adversarial training task involves blending the latent space of the encoder across English, Tulu, and Kannada, enabling the model to learn language-agnostic features~\citep{ko-etal-2021-adapting}. Without this task, while denoising autoencoding enhanced the robustness of the model's learned feature space, its benefits were somewhat compromised.

In the final step, Task 5, of the first iteration for the EN--TCY model, we performed additional fine-tuning using the EN--TCY sentence pairs utilized in Task 2. This process further increased the BLEU score of the model to 11.06 as the decoder adapted more effectively to the target (TCY) side. We used this model to generate Tulu sentences from the English side of the back-translated pairs. Subsequently, these new back-translated TCY-EN sentence pairs were utilized in the first iteration of training the TCY-EN model, the results of which are detailed in the following section.

\paragraph{Tulu-English Translation}

In Task 2 of the TCY--EN direction, the base TCY--EN model was trained using the back-translated TCY--EN pairs obtained from the previous step. This training resulted in a BLEU score of 19.53, indicating the effectiveness of transfer learning. The decoder demonstrated proficiency in generating English tokens, likely attributed to the pre-trained IndicBARTSS model's inherent language modeling capability in English.
Since the model had already been exposed to parallel KN--EN data in Task 1, the subsequent step involved denoising autoencoding with Kannada data (Task 4a). However, this led to a decline in the BLEU score to 7.08. Despite denoising autoencoding with Tulu data afterward, the BLEU score remained unchanged. The absence of the adversarial training task significantly limited the effectiveness of denoising autoencoding, as previously mentioned.
Nevertheless, in the subsequent task (Task 5), which involved fine-tuning with back-translated TCY--EN pairs for a second time, the BLEU score increased to 25.97. This represented the highest score attained by either model thus far.

\paragraph{Further Iterations}

We used the final TCY--EN model, which we obtained in Task 5, to subsequently back-translate the Tulu sentences that were used to train it in both Task 2 and Task 5. 
This newly generated set of back-translated EN--TCY pairs served to start a second iteration of the entire process. 
However, as shown in Table \ref{tab:results}, despite some initial improvement, the BLEU scores for the EN--TCY model kept declining from the starting score of 11.06.

\begin{table}[H]
\centering
\begin{tabular}{c|c|r|r}
\bf{Iter.} & \bf{Direction} & \bf{old BLEU} & \bf{new BLEU}    \\
\midrule
1   & EN--TCY   & 11.06 & 13.12 \\
1   & TCY--EN   & 25.97 & 21.85 \\
2   & EN--TCY   & 13.43 & 35.41 \\
\end{tabular}
\caption{BLEU scores after additional fine-tuning using \textit{DravidianLangTech-2022} data.}
\label{tab:additionalresults}
\end{table}

\paragraph{Fine-tuning with \textit{DravidianLangTech} data}
We used the parallel EN--TCY data from \textit{DravidianLangTech-22} to further fine-tune the EN--TCY and TCY--EN models obtained at the end of each iteration. Table \ref{tab:additionalresults} illustrates the changes in BLEU scores resulting from this fine-tuning step. 
For the EN--TCY model obtained at the end of Iteration 1, fine-tuning improved its BLEU score from 11.06 to 13.12. This moderate increase is not surprising, as the manually translated Tulu data would have further enhanced the decoder's performance. 
Conversely, for the TCY--EN model obtained at the end of Iteration 1, the BLEU score decreased from 25.97 to 21.85. The English sentences in this training data, generated by Google Translate, are not perfect translations and often contain transliterated Kannada words, particularly in the form of names of mythological characters, places, and local flora and fauna. We hypothesize that these aspects contributed to the degradation of the TCY--EN model's decoder. 
Finally, the EN--TCY model obtained at the end of Iteration 2 was also fine-tuned with this data, resulting in a substantial increase in its BLEU score from 13.43 to 35.41. 
This dramatic improvement may be attributed to the high-quality Tulu data, which eliminated spurious correlations in the latent space and simultaneously enhanced the decoder's ability to generate Tulu tokens.

\paragraph{Tulu Translation Performance}
The results suggest that the approach outlined by~\citet{ko-etal-2021-adapting} effectively achieves reasonable performance in translating from Tulu to English, as evidenced by the model's BLEU score of 25.97. To provide a point of comparison, we used Google Translate to translate the Tulu test data to English using Google Translate\footnote{\url{https://translate.google.com/}, in September 2023}. It automatically detected the sentences as Kannada and produced a translation with a BLEU score of 7.19.

However, the NMT-Adapt approach did not yield the same level of performance in the reverse direction, from English to Tulu. The highest achieved BLEU score, 17.27, was obtained through training with back-translation pairs and parallel English-Kannada data exclusively. However, when we combined the NMT-Adapt pipeline with additional fine-tuning using the \textit{DravidianLangTech-22} data, we observed a substantial improvement, resulting in a BLEU score of 35.41.

The denoising autoencoding step generated good results only when followed by fine-tuning with back-translation data in our setting. However, there are a multitude of factors, including the unique characters of Tulu and Kannada, as well as the constrained size of the monolingual Tulu dataset, consisting of just over 40,000 entries. 
In the work by~\citet{ko-etal-2021-adapting}, the monolingual datasets for all the `low-resource' languages they examined contained at least one million sentences. This implies that the model's decoder had a larger number of examples to adapt its feature space and acquire high-level semantic knowledge of the low-resource language.

\subsection{Qualitative Error Analysis}

To gain a qualitative understanding of the translations, we conducted an analysis by randomly selecting sentences from the model-generated translations and then comparing them with the reference translations. 
In the TCY--EN model with the highest BLEU score, we identified multiple cases where words were transliterated from Tulu to English rather than accurately translated.
Furthermore, we observed occurrences of Kannada characters appearing in English translations and, conversely, instances of English characters appearing in Kannada translations. 
Additionally, there were situations where common Tulu words, which had distinct meanings in Kannada or closely resembled common Kannada words, were translated as Kannada instead of Tulu.

For instance, the Tulu word \texttt{uppuna}, which should have been translated as \textit{together}, was incorrectly translated to \textit{salt}, which is \texttt{uppu} in Kannada. 
Similarly, the phrase \texttt{tenkāyi amērikā}, which means \textit{South America} in Tulu, was translated into English as \textit{United States} by the TCY--EN model, ignoring the first word. However, \texttt{dakṣiṇa āphrikā}, which is the formal name for South Africa in both Tulu and Kannada, was correctly translated as \textit{South Africa}. This discrepancy arises from the fact that \texttt{dakṣiṇa} is a loan word from Sanskrit for \textit{south} used in both Kannada and Tulu, whereas \texttt{tenkāyi} is unique to Tulu. 

Finally, we observed instances of word repetition or recurring sequences of words in the translations known as hallucinations. This phenomenon is a well-documented challenge in text generation tasks, as discussed in~\citep{Fu_Lam_So_Shi_2021}.

\section{Conclusion}

We introduced the first parallel dataset for English--Tulu by incorporating Tulu translations into the multilingual machine translation resource FLORES-200~\citeplanguageresource{flores,nllb-22}.

Furthermore, we developed a machine translation system for English--Tulu by leveraging resources for Kannada, a related South Dravidian language. We employed a transfer learning approach that exploits the similarities between the languages, enabling the training of a machine translation system even in the absence of parallel data between the source and target languages.

Our system achieved a BLEU score of 25.97 for Tulu--English translation, significantly outperforming Google Translate in September 2023, which reached a  BLEU of 7.19.
However, the relatively low BLEU scores indicate that the usefulness of our system's translations is limited. In English--Tulu translation, the model often retains elements of Kannada in the output. However, in Tulu--English translation, we observe that certain parts of Tulu sentences in the test set are conveyed effectively enough for non-Tulu speakers to understand. Additionally, proper nouns are accurately transliterated into English in the results. However, the translation quality diminishes as sentences become longer, and in some cases, the model simply transliterates complex Tulu words into English.

\section{Limitations}\label{limitations}

\citet{ko-etal-2021-adapting} implemented NMT-Adapt using the fairseq toolkit\footnote{\url{https://github.com/facebookresearch/fairseq}}~\citep{ott2019fairseq} and mBART~\cite{mbart} as the pre-trained model. However, since Kannada is not supported in mBART, we worked with the pre-trained IndicBARTSS model and the YANMTT toolkit. Unfortunately, YANMTT does not include the adversarial training with Wasserstein loss, a critical step in NMT-Adapt for achieving the objectives of denoising autoencoding.
In future work, we plan to implement this step and integrate it into the toolkit.

Furthermore, to benchmark against NMT-Adapt models, we attempted to train bilingual (EN--KN) and trilingual (EN--KN--ML) transformer models with the intention of subsequently adapting them to translate Tulu. 
Nevertheless, due to resource constraints, particularly when initialized with sizes akin to mBART or IndicBARTSS, these models proved too large to train. Smaller models trained with the Samanantar dataset achieved a maximum BLEU score of only 8.60 when translating EN--KN.

Finally, we used the parallel EN--TCY data adapted from \textit{DravidianLangTech-22} to independently fine-tune models at the end of each iteration. 
To ensure that improvements are consistently incorporated into each subsequent iteration, this step should be incorporated into the NMT-Adapt pipeline.
This would be important to gain a more comprehensive understanding of this step and potentially quantify its effects.

\section{Ethics Statement}
This paper introduces a novel dataset for Tulu and presents initial efforts towards developing a machine translation system for English--Tulu. 
We collaborated with \textit{Jai Tulunad}, a volunteer organization based in India, dedicated to preserving Tulu language and culture. Their enthusiastic contribution to our project assures us that our efforts align with the community's interests and needs.

\section{Acknowledgements}

We thank Chantal Amrhein for her support, Anne Göhring, Amit Moryossef, and the anonymous reviewers for their insightful input. Additionally, we extend our appreciation to the organization \textit{Jai Tulunad} and the Tulu translators for their invaluable contributions to this study. This work was supported by the Swiss National Science Foundation (project no. 191934) and the Department of Computational Linguistics at the University of Zurich.

\newpage

\section{Bibliographical References}\label{sec:reference}

\bibliographystyle{lrec_natbib}
\bibliography{lrec-coling2024-example,acl_anthology_part1,custom} 

\begin{thebibliography}{4}
\expandafter\ifx\csname natexlab\endcsname\relax\def\natexlab#1{#1}\fi

\bibitem[{Goyal et~al.(2022)Goyal, Gao, Chaudhary, Chen, Wenzek, Ju, Krishnan, Ranzato, Guzm{\'a}n, and Fan}]{flores}
Naman Goyal, Cynthia Gao, Vishrav Chaudhary, Peng-Jen Chen, Guillaume Wenzek, Da~Ju, Sanjana Krishnan, Marc{'}Aurelio Ranzato, Francisco Guzm{\'a}n, and Angela Fan. 2022.
\newblock \href {https://doi.org/10.1162/tacl_a_00474} {The {F}lores-101 evaluation benchmark for low-resource and multilingual machine translation}.
\newblock \emph{Transactions of the Association for Computational Linguistics}, 10:522--538.

\bibitem[{Madasamy et~al.(2022)Madasamy, Hegde, Banerjee, Chakravarthi, Priyadharshini, Shashirekha, and McCrae}]{madasamy-etal-2022-overview-this}
Anand~Kumar Madasamy, Asha Hegde, Shubhanker Banerjee, Bharathi~Raja Chakravarthi, Ruba Priyadharshini, Hosahalli Shashirekha, and John McCrae. 2022.
\newblock \href {https://doi.org/10.18653/v1/2022.dravidianlangtech-1.41} {Overview of the shared task on machine translation in {D}ravidian languages}.
\newblock In \emph{Proceedings of the Second Workshop on Speech and Language Technologies for Dravidian Languages}, pages 271--278, Dublin, Ireland. Association for Computational Linguistics.

\bibitem[{{NLLB Team} et~al.(2022){NLLB Team}, Costa-jussà, Cross, Çelebi, Elbayad, Heafield, Heffernan, Kalbassi, Lam, Licht, Maillard, Sun, Wang, Wenzek, Youngblood, Akula, Barrault, Gonzalez, Hansanti, Hoffman, Jarrett, Sadagopan, Rowe, Spruit, Tran, Andrews, Ayan, Bhosale, Edunov, Fan, Gao, Goswami, Guzmán, Koehn, Mourachko, Ropers, Saleem, Schwenk, and Wang}]{nllb-22}
{NLLB Team}, Marta~R. Costa-jussà, James Cross, Onur Çelebi, Maha Elbayad, Kenneth Heafield, Kevin Heffernan, Elahe Kalbassi, Janice Lam, Daniel Licht, Jean Maillard, Anna Sun, Skyler Wang, Guillaume Wenzek, Al~Youngblood, Bapi Akula, Loic Barrault, Gabriel~Mejia Gonzalez, Prangthip Hansanti, John Hoffman, Semarley Jarrett, Kaushik~Ram Sadagopan, Dirk Rowe, Shannon Spruit, Chau Tran, Pierre Andrews, Necip~Fazil Ayan, Shruti Bhosale, Sergey Edunov, Angela Fan, Cynthia Gao, Vedanuj Goswami, Francisco Guzmán, Philipp Koehn, Alexandre Mourachko, Christophe Ropers, Safiyyah Saleem, Holger Schwenk, and Jeff Wang. 2022.
\newblock \href {http://arxiv.org/abs/2207.04672} {No language left behind: Scaling human-centered machine translation}.

\bibitem[{Ramesh et~al.(2022)Ramesh, Doddapaneni, Bheemaraj, Jobanputra, AK, Sharma, Sahoo, Diddee, J, Kakwani, Kumar, Pradeep, Nagaraj, Deepak, Raghavan, Kunchukuttan, Kumar, and Khapra}]{Samanantar}
Gowtham Ramesh, Sumanth Doddapaneni, Aravinth Bheemaraj, Mayank Jobanputra, Raghavan AK, Ajitesh Sharma, Sujit Sahoo, Harshita Diddee, Mahalakshmi J, Divyanshu Kakwani, Navneet Kumar, Aswin Pradeep, Srihari Nagaraj, Kumar Deepak, Vivek Raghavan, Anoop Kunchukuttan, Pratyush Kumar, and Mitesh~Shantadevi Khapra. 2022.
\newblock \href {https://doi.org/10.1162/tacl_a_00452} {Samanantar: The largest publicly available parallel corpora collection for 11 {I}ndic languages}.
\newblock \emph{Transactions of the Association for Computational Linguistics}, 10:145--162.

\end{thebibliography}


\begin{thebibliography}{37}
\expandafter\ifx\csname natexlab\endcsname\relax\def\natexlab#1{#1}\fi

\bibitem[{Artetxe et~al.(2018)Artetxe, Labaka, Agirre, and Cho}]{artetxe2018unsupervised}
Mikel Artetxe, Gorka Labaka, Eneko Agirre, and Kyunghyun Cho. 2018.
\newblock \href {https://openreview.net/forum?id=Sy2ogebAW} {Unsupervised neural machine translation}.
\newblock In \emph{International Conference on Learning Representations}.

\bibitem[{Attardi(2015)}]{Wikiextractor2015}
Giuseppe Attardi. 2015.
\newblock {WikiExtractor}.
\newblock \url{https://github.com/attardi/wikiextractor}.

\bibitem[{Bahdanau et~al.(2015)Bahdanau, Cho, and Bengio}]{bahdanau}
Dzmitry Bahdanau, Kyunghyun Cho, and Yoshua Bengio. 2015.
\newblock \href {http://arxiv.org/abs/1409.0473} {Neural machine translation by jointly learning to align and translate}.
\newblock In \emph{3rd International Conference on Learning Representations, {ICLR} 2015, San Diego, CA, USA, May 7-9, 2015, Conference Track Proceedings}.

\bibitem[{Bala~Das et~al.(2023)Bala~Das, Biradar, Kumar~Mishra, and Kr.~Patra}]{10.1145/3587932}
Sudhansu Bala~Das, Atharv Biradar, Tapas Kumar~Mishra, and Bidyut Kr.~Patra. 2023.
\newblock \href {https://doi.org/10.1145/3587932} {Improving multilingual neural machine translation system for indic languages}.
\newblock \emph{ACM Trans. Asian Low-Resour. Lang. Inf. Process.}, 22(6).

\bibitem[{Bird(2020)}]{bird-2020-decolonising}
Steven Bird. 2020.
\newblock \href {https://doi.org/10.18653/v1/2020.coling-main.313} {Decolonising speech and language technology}.
\newblock In \emph{Proceedings of the 28th International Conference on Computational Linguistics}, pages 3504--3519, Barcelona, Spain (Online). International Committee on Computational Linguistics.

\bibitem[{Bird(2022)}]{bird-2022-local}
Steven Bird. 2022.
\newblock \href {https://doi.org/10.18653/v1/2022.acl-long.539} {Local languages, third spaces, and other high-resource scenarios}.
\newblock In \emph{Proceedings of the 60th Annual Meeting of the Association for Computational Linguistics (Volume 1: Long Papers)}, pages 7817--7829, Dublin, Ireland. Association for Computational Linguistics.

\bibitem[{Blaschke et~al.(2024)Blaschke, Purschke, Schütze, and Plank}]{blaschke2024dialect}
Verena Blaschke, Christoph Purschke, Hinrich Schütze, and Barbara Plank. 2024.
\newblock \href {http://arxiv.org/abs/2402.11968} {What do dialect speakers want? a survey of attitudes towards language technology for german dialects}.

\bibitem[{Brigel(1982)}]{brigel1982grammar}
J.~Brigel. 1982.
\newblock \href {https://books.google.ch/books?id=81AgB3u2tcQC} {\emph{A Grammar of the Tulu Language}}.
\newblock Asian Educational Services.

\bibitem[{Dabre et~al.(2023)Dabre, Kanojia, Sawant, and Sumita}]{dabre-etal-2023-yanmtt}
Raj Dabre, Diptesh Kanojia, Chinmay Sawant, and Eiichiro Sumita. 2023.
\newblock \href {https://doi.org/10.18653/v1/2023.acl-demo.24} {{YANMTT}: Yet another neural machine translation toolkit}.
\newblock In \emph{Proceedings of the 61st Annual Meeting of the Association for Computational Linguistics (Volume 3: System Demonstrations)}, pages 257--263, Toronto, Canada. Association for Computational Linguistics.

\bibitem[{Dabre et~al.(2022)Dabre, Shrotriya, Kunchukuttan, Puduppully, Khapra, and Kumar}]{indicbart}
Raj Dabre, Himani Shrotriya, Anoop Kunchukuttan, Ratish Puduppully, Mitesh Khapra, and Pratyush Kumar. 2022.
\newblock \href {https://doi.org/10.18653/v1/2022.findings-acl.145} {{I}ndic{BART}: A pre-trained model for indic natural language generation}.
\newblock In \emph{Findings of the Association for Computational Linguistics: ACL 2022}, pages 1849--1863, Dublin, Ireland. Association for Computational Linguistics.

\bibitem[{Eberhard et~al.(2023)Eberhard, Simons, and Fennig}]{ethnologue}
David~M. Eberhard, Gary~F. Simons, and Charles~D. Fennig, editors. 2023.
\newblock \emph{Ethnologue: Languages of the World}, twenty-sixth edition.
\newblock SIL International, Dallas, Texas.
\newblock Online version: \url{http://www.ethnologue.com}.

\bibitem[{Fu et~al.(2021)Fu, Lam, So, and Shi}]{Fu_Lam_So_Shi_2021}
Zihao Fu, Wai Lam, Anthony Man-Cho So, and Bei Shi. 2021.
\newblock \href {https://doi.org/10.1609/aaai.v35i14.17520} {A theoretical analysis of the repetition problem in text generation}.
\newblock \emph{Proceedings of the AAAI Conference on Artificial Intelligence}, 35(14):12848--12856.

\bibitem[{Goyal et~al.(2022)Goyal, Supriya, U, and Nayak}]{goyal-etal-2022-translation}
Piyushi Goyal, Musica Supriya, Dinesh U, and Ashalatha Nayak. 2022.
\newblock \href {https://doi.org/10.18653/v1/2022.dravidianlangtech-1.19} {Translation techies @{D}ravidian{L}ang{T}ech-{ACL}2022-machine translation in {D}ravidian languages}.
\newblock In \emph{Proceedings of the Second Workshop on Speech and Language Technologies for Dravidian Languages}, pages 120--124, Dublin, Ireland. Association for Computational Linguistics.

\bibitem[{Kakwani et~al.(2020)Kakwani, Kunchukuttan, Golla, N.C., Bhattacharyya, Khapra, and Kumar}]{indicnlpsuite}
Divyanshu Kakwani, Anoop Kunchukuttan, Satish Golla, Gokul N.C., Avik Bhattacharyya, Mitesh~M. Khapra, and Pratyush Kumar. 2020.
\newblock \href {https://doi.org/10.18653/v1/2020.findings-emnlp.445} {{I}ndic{NLPS}uite: Monolingual corpora, evaluation benchmarks and pre-trained multilingual language models for {I}ndian languages}.
\newblock In \emph{Findings of the Association for Computational Linguistics: EMNLP 2020}, pages 4948--4961, Online. Association for Computational Linguistics.

\bibitem[{Klein et~al.(2017)Klein, Kim, Deng, Senellart, and Rush}]{klein-etal-2017-opennmt}
Guillaume Klein, Yoon Kim, Yuntian Deng, Jean Senellart, and Alexander Rush. 2017.
\newblock \href {https://aclanthology.org/P17-4012} {{O}pen{NMT}: Open-source toolkit for neural machine translation}.
\newblock In \emph{Proceedings of {ACL} 2017, System Demonstrations}, pages 67--72, Vancouver, Canada. Association for Computational Linguistics.

\bibitem[{Ko et~al.(2021)Ko, El-Kishky, Renduchintala, Chaudhary, Goyal, Guzm{\'a}n, Fung, Koehn, and Diab}]{ko-etal-2021-adapting}
Wei-Jen Ko, Ahmed El-Kishky, Adithya Renduchintala, Vishrav Chaudhary, Naman Goyal, Francisco Guzm{\'a}n, Pascale Fung, Philipp Koehn, and Mona Diab. 2021.
\newblock \href {https://doi.org/10.18653/v1/2021.acl-long.66} {Adapting high-resource {NMT} models to translate low-resource related languages without parallel data}.
\newblock In \emph{Proceedings of the 59th Annual Meeting of the Association for Computational Linguistics and the 11th International Joint Conference on Natural Language Processing (Volume 1: Long Papers)}, pages 802--812, Online. Association for Computational Linguistics.

\bibitem[{Koehn and Knowles(2017)}]{6challenges}
Philipp Koehn and Rebecca Knowles. 2017.
\newblock \href {https://doi.org/10.18653/v1/W17-3204} {Six challenges for neural machine translation}.
\newblock In \emph{Proceedings of the First Workshop on Neural Machine Translation}, pages 28--39, Vancouver. Association for Computational Linguistics.

\bibitem[{Krishnamurti(2003)}]{krishnamurti_2003}
Bhadriraju Krishnamurti. 2003.
\newblock \href {https://doi.org/10.1017/CBO9780511486876} {\emph{The Dravidian Languages}}.
\newblock Cambridge Language Surveys. Cambridge University Press.

\bibitem[{Lakew et~al.(2020)Lakew, Negri, and Turchi}]{Lakew2020LowRN}
Surafel~Melaku Lakew, Matteo Negri, and Marco Turchi. 2020.
\newblock \href {https://api.semanticscholar.org/CorpusID:214727851} {Low resource neural machine translation: A benchmark for five african languages}.
\newblock \emph{ArXiv}, abs/2003.14402.

\bibitem[{Lample et~al.(2018)Lample, Conneau, Denoyer, and Ranzato}]{lample2018unsupervised}
Guillaume Lample, Alexis Conneau, Ludovic Denoyer, and Marc'Aurelio Ranzato. 2018.
\newblock \href {https://openreview.net/forum?id=rkYTTf-AZ} {Unsupervised machine translation using monolingual corpora only}.
\newblock In \emph{International Conference on Learning Representations}.

\bibitem[{Littauer and Paterson~III(2016)}]{littauer2016open}
Richard Littauer and Hugh Paterson~III. 2016.
\newblock Open source code serving endangered languages.
\newblock In \emph{Proceedings of LREC 2016 Collaboration and Computing for Under-Resourced Languages: Towards an Alliance for Digital Language Diversity (CCURL) Workshop}, pages 86--88, Portorož, Slovenia.

\bibitem[{Liu et~al.(2020)Liu, Gu, Goyal, Li, Edunov, Ghazvininejad, Lewis, and Zettlemoyer}]{mbart}
Yinhan Liu, Jiatao Gu, Naman Goyal, Xian Li, Sergey Edunov, Marjan Ghazvininejad, Mike Lewis, and Luke Zettlemoyer. 2020.
\newblock \href {https://doi.org/10.1162/tacl_a_00343} {Multilingual denoising pre-training for neural machine translation}.
\newblock \emph{Transactions of the Association for Computational Linguistics}, 8:726--742.

\bibitem[{Liu et~al.(2022)Liu, Richardson, Hatcher, and Prud{'}hommeaux}]{liu-etal-2022-always}
Zoey Liu, Crystal Richardson, Richard Hatcher, and Emily Prud{'}hommeaux. 2022.
\newblock \href {https://doi.org/10.18653/v1/2022.acl-long.272} {Not always about you: Prioritizing community needs when developing endangered language technology}.
\newblock In \emph{Proceedings of the 60th Annual Meeting of the Association for Computational Linguistics (Volume 1: Long Papers)}, pages 3933--3944, Dublin, Ireland. Association for Computational Linguistics.

\bibitem[{Mukhija et~al.(2021)Mukhija, Choudhury, and Bali}]{mukhija2021designing}
Namrata Mukhija, Monojit Choudhury, and Kalika Bali. 2021.
\newblock \href {http://arxiv.org/abs/2110.07444} {Designing language technologies for social good: The road not taken}.

\bibitem[{{Office of the Registrar General \& Census Commissioner, India}(2022)}]{language_atlas_kan}
{Office of the Registrar General \& Census Commissioner, India}. 2022.
\newblock \emph{Distribution of Kannada Speakers 2011}, pages 38--39. Government of India.

\bibitem[{Ott et~al.(2019)Ott, Edunov, Baevski, Fan, Gross, Ng, Grangier, and Auli}]{ott2019fairseq}
Myle Ott, Sergey Edunov, Alexei Baevski, Angela Fan, Sam Gross, Nathan Ng, David Grangier, and Michael Auli. 2019.
\newblock fairseq: A fast, extensible toolkit for sequence modeling.
\newblock In \emph{Proceedings of NAACL-HLT 2019: Demonstrations}.

\bibitem[{Papineni et~al.(2002)Papineni, Roukos, Ward, and Zhu}]{papineni-etal-2002-bleu}
Kishore Papineni, Salim Roukos, Todd Ward, and Wei-Jing Zhu. 2002.
\newblock \href {https://doi.org/10.3115/1073083.1073135} {{B}leu: a method for automatic evaluation of machine translation}.
\newblock In \emph{Proceedings of the 40th Annual Meeting of the Association for Computational Linguistics}, pages 311--318, Philadelphia, Pennsylvania, USA. Association for Computational Linguistics.

\bibitem[{Post(2018)}]{sacrebleu}
Matt Post. 2018.
\newblock \href {https://doi.org/10.18653/v1/W18-6319} {A call for clarity in reporting {BLEU} scores}.
\newblock In \emph{Proceedings of the Third Conference on Machine Translation: Research Papers}, pages 186--191, Brussels, Belgium. Association for Computational Linguistics.

\bibitem[{Sennrich et~al.(2016)Sennrich, Haddow, and Birch}]{sennrich-etal-2016-improving}
Rico Sennrich, Barry Haddow, and Alexandra Birch. 2016.
\newblock \href {https://doi.org/10.18653/v1/P16-1009} {Improving neural machine translation models with monolingual data}.
\newblock In \emph{Proceedings of the 54th Annual Meeting of the Association for Computational Linguistics (Volume 1: Long Papers)}, pages 86--96, Berlin, Germany. Association for Computational Linguistics.

\bibitem[{Steever(2017)}]{steever_2017}
Sanford~B. Steever. 2017.
\newblock \href {https://doi.org/10.1017/9781316135716.028} {The dravidian language family}.
\newblock In \emph{The Cambridge Handbook of Linguistic Typology}, Cambridge Handbooks in Language and Linguistics, page 887–910. Cambridge University Press.

\bibitem[{Steever(2019)}]{steever2019dravidian}
S.B. Steever, editor. 2019.
\newblock \href {https://doi.org/10.4324/9781315722580} {\emph{The Dravidian Languages}}, 2nd edition.
\newblock Routledge.

\bibitem[{Subrahmanyam(2006)}]{SUBRAHMANYAM2006785}
P.S. Subrahmanyam. 2006.
\newblock \href {https://doi.org/https://doi.org/10.1016/B0-08-044854-2/02147-7} {Dravidian languages}.
\newblock In Keith Brown, editor, \emph{Encyclopedia of Language \& Linguistics (Second Edition)}, second edition edition, pages 785--795. Elsevier, Oxford.

\bibitem[{Sutskever et~al.(2014)Sutskever, Vinyals, and Le}]{seq2seq}
Ilya Sutskever, Oriol Vinyals, and Quoc~V Le. 2014.
\newblock \href {https://proceedings.neurips.cc/paper_files/paper/2014/file/a14ac55a4f27472c5d894ec1c3c743d2-Paper.pdf} {Sequence to sequence learning with neural networks}.
\newblock In \emph{Advances in Neural Information Processing Systems}, volume~27. Curran Associates, Inc.

\bibitem[{Thadhagath(2023)}]{hindustan_times_2023}
Pathi~Venkata Thadhagath. 2023.
\newblock \href {https://www.hindustantimes.com/cities/bengaluru-news/demand-to-make-tulu-second-official-language-of-karnataka-arises-in-assembly-101689751745948.html} {Demand to make tulu second official language of karnataka arises in assembly}.
\newblock Hindustan Times.

\bibitem[{Vaswani et~al.(2017)Vaswani, Shazeer, Parmar, Uszkoreit, Jones, Gomez, Kaiser, and Polosukhin}]{attention}
Ashish Vaswani, Noam Shazeer, Niki Parmar, Jakob Uszkoreit, Llion Jones, Aidan~N Gomez, \L~ukasz Kaiser, and Illia Polosukhin. 2017.
\newblock \href {https://proceedings.neurips.cc/paper_files/paper/2017/file/3f5ee243547dee91fbd053c1c4a845aa-Paper.pdf} {Attention is all you need}.
\newblock In \emph{Advances in Neural Information Processing Systems}, volume~30. Curran Associates, Inc.

\bibitem[{Wikipedia(2023)}]{Tulu_Wikipedia}
Wikipedia. 2023.
\newblock {Tulu Wikipedia} --- {W}ikipedia{,} the free encyclopedia.
\newblock \url{http://en.wikipedia.org/w/index.php?title=Tulu\%20Wikipedia&oldid=1157462605}.
\newblock [Online; accessed 29-May-2023].

\bibitem[{Zoph et~al.(2016)Zoph, Yuret, May, and Knight}]{zoph-etal-2016-transfer}
Barret Zoph, Deniz Yuret, Jonathan May, and Kevin Knight. 2016.
\newblock \href {https://doi.org/10.18653/v1/D16-1163} {Transfer learning for low-resource neural machine translation}.
\newblock In \emph{Proceedings of the 2016 Conference on Empirical Methods in Natural Language Processing}, pages 1568--1575, Austin, Texas. Association for Computational Linguistics.

\end{thebibliography}

\section{Language Resource References}
\label{lr:ref}

\bibliographystylelanguageresource{lrec_natbib}
\bibliographylanguageresource{languageresource}

\appendix

\end{document}